\documentclass[10pt,twocolumn,letterpaper]{article}

\usepackage[pagenumbers]{cvpr} % To force page numbers, e.g. for an arXiv version

\usepackage{graphicx}
\usepackage{amsmath}
\usepackage{amssymb}
\usepackage{booktabs}
\usepackage{tabularx}

\usepackage[accsupp]{axessibility}
\usepackage{array}
\usepackage{adjustbox}
\usepackage{stfloats}
\usepackage{array}
\usepackage{bbm}
\usepackage{float}
\usepackage{longtable}
\usepackage{subcaption}

\usepackage[pagebackref,breaklinks,colorlinks]{hyperref}

\usepackage[capitalize]{cleveref}
\crefname{section}{Sec.}{Secs.}
\Crefname{section}{Section}{Sections}
\Crefname{table}{Table}{Tables}
\crefname{table}{Tab.}{Tabs.}

\def\cvprPaperID{8} % *** Enter the CVPR Paper ID here
\def\confName{CVPR}
\def\confYear{2022}

\usepackage{overpic}
\usepackage{enumitem} %< control spacing in itemize/enumerate/...
\usepackage{overpic} %< add raw math symbols to figures
\usepackage{color}
\usepackage{array}
\usepackage{adjustbox}
\usepackage{stfloats}
\usepackage{array}
\usepackage{tabularx}
\usepackage{float}
\usepackage{longtable}

\definecolor{turquoise}{cmyk}{0.65,0,0.1,0.3}
\definecolor{purple}{rgb}{0.65,0,0.65}
\definecolor{dark_green}{rgb}{0, 0.5, 0}
\definecolor{orange}{rgb}{0.8, 0.6, 0.2}
\definecolor{red}{rgb}{0.8, 0.2, 0.2}
\definecolor{darkred}{rgb}{0.6, 0.1, 0.05}
\definecolor{blueish}{rgb}{0.0, 0.3, .6}
\definecolor{light_gray}{rgb}{0.7, 0.7, .7}
\definecolor{pink}{rgb}{1, 0, 1}
\definecolor{greyblue}{rgb}{0.25, 0.25, 1}

\usepackage{blindtext}

\renewcommand{\paragraph}[1]{\noindent\textbf{#1}.}
\begin{document}

%%%%%%%%% TITLE - PLEASE UPDATE
\title{Adversarial Robustness through the Lens of Convolutional Filters}

\author{Paul Gavrikov$^1$%
\thanks{Funded by the Ministry for Science, Research and Arts, Baden-Wuerttemberg, Grant 32-7545.20/45/1 (Q-AMeLiA).}
\ and Janis Keuper$^{1,2,3}$\footnotemark[1]\\
$^1$IMLA, Offenburg University, $^2$CC-HPC, Fraunhofer ITWM, $^3$Fraunhofer Research Center ML\\
{\tt\small \{first.last\}@hs-offenburg.de}
}
\maketitle

%%%%%%%%% ABSTRACT
\begin{abstract}
Deep learning models are intrinsically sensitive to distribution shifts in the input data. In particular, small, barely perceivable perturbations to the input data can force models to make wrong predictions with high confidence. An common defense mechanism is regularization through adversarial training which injects worst-case perturbations back into training to strengthen the decision boundaries, and to reduce overfitting. In this context, we perform an investigation of $3\times 3$ convolution filters that form in adversarially-trained models. Filters are extracted from 71 public models of the $\ell_\infty$-RobustBench \textit{CIFAR-10/100} and \textit{ImageNet1k} leaderboard and compared to filters extracted from models built on the same architectures but trained without robust regularization. We observe that adversarially-robust models appear to form more diverse, less sparse, and more orthogonal convolution filters than their normal counterparts. The largest differences between robust and normal models are found in the deepest layers, and the very first convolution layer, which consistently and predominantly forms filters that can partially eliminate perturbations, irrespective of the architecture.

{\noindent\normalfont\textbf{Data \& Project website:}\\
\url{https://github.com/paulgavrikov/cvpr22w_RobustnessThroughTheLens}}
\end{abstract}

%%%%%%%%% BODY TEXT
\section{Introduction}\label{sec:intro}
Convolutional Neural Networks (CNNs) have been successfully applied to solve many different computer vision problems. As the state of the art has been consequently pushed, research was mostly devoted at improving the performance (validation accuracy, along speed and others). However, recently it has been shown that these models are sensitive to distribution shifts in image data. Even small, for humans almost imperceptible, perturbations applied to input images can force the networks to make high-confidence, false predictions on samples that would otherwise have been classified correctly\cite{Szegedy2013Dec,Moosavi-Dezfooli2015Nov}. Normal training on off-the-shelf architectures typically results in zero validation accuracy against perturbed samples. This raises the question on whether current deep learning models should be used in safety-critical applications \cite{MA2021107332,doi:10.1126/science.aaw4399, deng2020analysis}. Consequently, researchers have devoted their work on studying the sensitivity to distribution shifts \eg by finding and understanding adversarial inputs \cite{Carlini2016Aug,akhtar2018threat}, and building defenses to those \cite{Papernot2015Nov,Goodfellow2014Dec,madry2018towards,Shafahi2019Apr}. 
While most explanatory methods study the distribution shifts in the input data and activations, we propose to evaluate differences in learned convolutional filters and, therefore, round out previous findings through a different perspective. More specifically, we investigate shifts in the dominantly used $3\times 3$ filters in CNN classification models trained on \textit{CIFAR-10/100}\cite{cifar} and \textit{ImageNet1k}\cite{imnet} datasets that were trained to withstand $\ell_\infty$-bound adversarial attacks. However, we believe that our results also apply to other tasks, datasets, and perhaps even to other attack vectors. 
We summarize our key contributions and findings as follows:
\begin{figure}
  \centering
  \includegraphics[width=\linewidth]{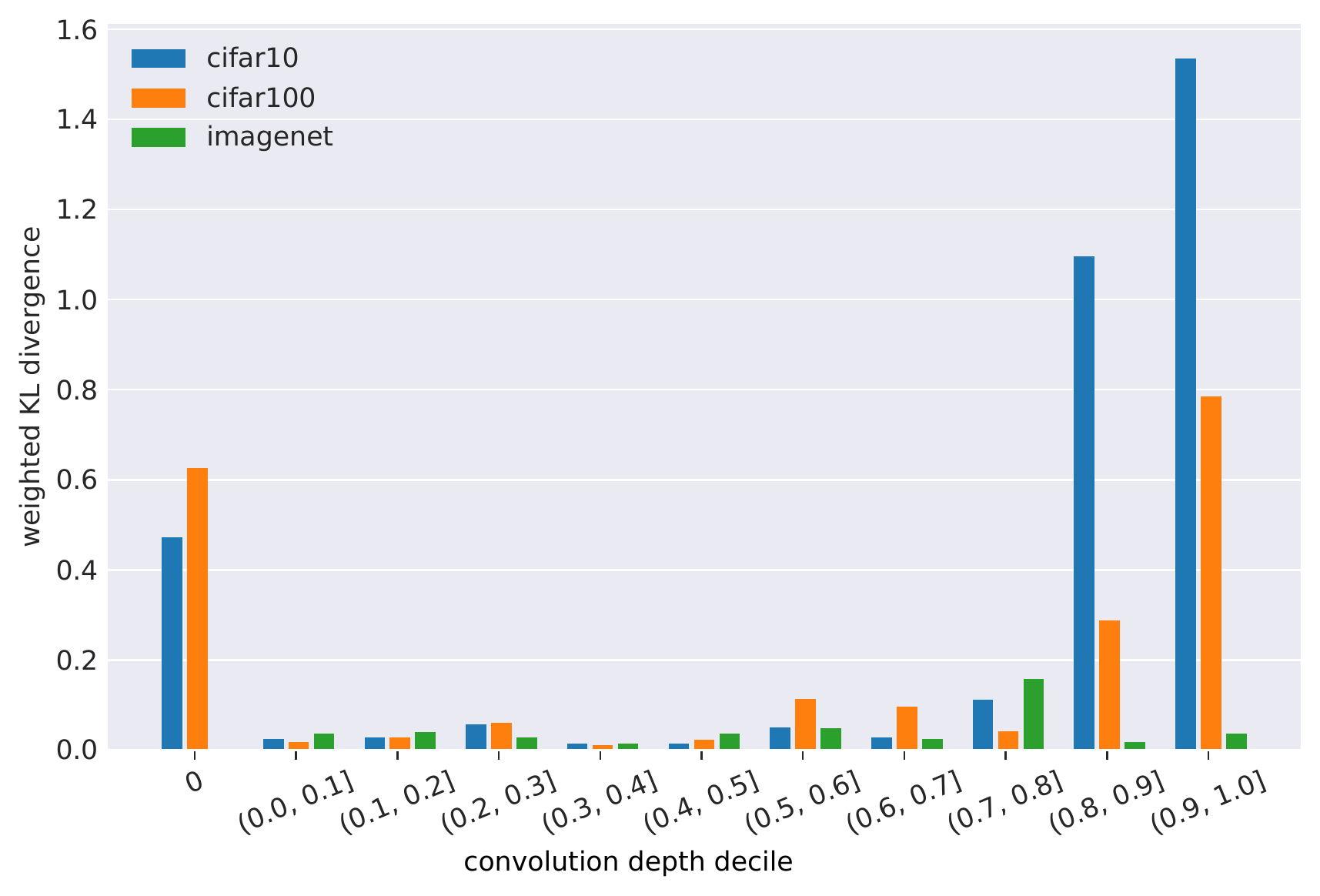}
  \caption{Filter structure divergence between learned $3\times 3$ filter structures of robust and normal models by depth decile. The first convolutional layer is displayed separately. The most significant shifts (large KL values) appear in the primary convolution layer and deeper stages.}
  \label{fig:divergence_by_depth}
\end{figure}

\begin{itemize}[leftmargin=*]
\setlength\itemsep{-.3em}
\item We collect 71 public robust models with 13 different architectures trained on 3 image datasets. These models contain a total of 615,863,744 filters with a size of $3\times 3$. Additionally, the therein used architectures are trained from scratch without the employed robustness regularizations.
\item We show an in depth empirical comparison of learned $3\times~3$ convolution filters between robust and normal models. The resulting filter dataset is made available publicly\cite{cnnfilterdb_robust}.
\item Our analysis shows that differences in filter structure increase with layer depth, but significantly explode towards the end of the model, with a dominant outlier showing in the primary convolution layer.
\item We visualize the primary layer of robust models and its activations, and observe a large presence of  thresholding-filters that can remove perturbations from regions of interest.  
\item We discover that robust models appear to form more diverse, less sparse, and more orthogonal convolution filters.
\end{itemize}
%
%
%-------------------------------------------------------------------------
%
\section{Related Work}\label{sec:related_work}
\paragraph{Adversarial attacks and defenses}
Let $\mathcal{F}$ denote a model parameterized by $\theta$, $x$ an input sample with the corresponding class label $y$, and $\mathcal{L}$ the loss function. Adversarial attacks attempt to maximize the loss $\mathcal{L}$ by finding an additive perturbation to an input sample $x'$ in the $\mathcal{B}_{\epsilon}(x)$ ball that is centered at $x$ with a radius of $\epsilon$. $\|\cdot\|_p$ depicts the $\ell_p$-norm with usually $p=2$ or $\infty$.
\begin{equation}
    \begin{gathered}
        \displaystyle\max_{x'\in \mathcal{B}_{\epsilon}(x)}\mathcal{L}\left(\mathcal{F}\left(x'; \theta\right), y\right)\\
        \mathcal{B}_{\epsilon}(x) = \{x':\|x-x'\|_{p}\leq\epsilon\}
    \end{gathered}
\end{equation}
On the other hand, to achieve robustness, the loss caused by the perturbation has to be mitigated by finding more suitable set of model parameters $\theta$. One of the most successful approaches is seen in \textit{adversarial training}\cite{madry2018towards} where adversarial perturbations are found and reintroduced into training, alongside with the inclusion of external data \cite{carmon2022unlabeled}.\\

\paragraph{Robustness evaluation} A common framework for adversarial robustness benchmarks is \textit{RobustBench} \cite{croce2021robustbench}. The framework applies $\text{APGD}_{ce}$, $\text{APGD}_{t}$ \cite{madry2018towards, croce2020reliable}, \textit{FAB} \cite{croce2020minimally}, and \textit{Square} \cite{10.1007/978-3-030-58592-1_29} attacks via \textit{AutoAttack} \cite{croce2020reliable} to obtain a comparable robustness accuracy. Perturbations are obtained from $\mathcal{B}_{\epsilon}$ with $p=2, \epsilon = 0.5$ on \textit{CIFAR-10}, as well as $p=\infty, \epsilon=8/255$ on \textit{CIFAR-10/100}, and $p=\infty, \epsilon=4/255$ on \textit{ImageNet1k}, respectively.
This methodology was questioned more recently, as the established $\epsilon$-thresholds are disputed as being too large and generate perturbations that can easily be detected\cite{lorenz2021detecting}.\\

\paragraph{Filter analysis}
\textit{Yosinski et. al.} \cite{Yosinski2014} studied filters of \textit{ImageNet1k} CNN classification models and concluded that early vision layers will form similar features, namely Gabor-filters and color-blobs, independent of task or dataset. On the other hand, deeper layers will capture specifics of the dataset by forming increasingly specialized filters.
A thorough analysis of filters limited to a specific \textit{InceptionV1} \cite{inception} model trained on \textit{ImageNet1k} was presented in \cite{Olah2020,olah2020an,cammarata2020curve,olah2020naturally,schubert2021high-low,cammarata2021curve,voss2021visualizing,voss2021branch,petrov2021weight}. The authors back \textit{Yosinski et. al.} and even go beyond by arguing that models are not only forming similar filters, but also connections (i.e. consecutive transformations). As model capacity increased, little to no research was devoted to understanding learned filters. More recently, we presented an empirical analysis of 1.4B filters obtained from models with different architectures, datasets, and training tasks \cite{anonCVPRPaper}. We also introduced a PCA-based method to compare the structure of learned filters, alongside two metrics to evaluate their quality (\textit{sparsity} and \textit{variance entropy}). Our findings showed that learned filter distributions remain largely similar across various splits, but many models show a large ration of degradation in filters. Within our study, we also briefly touched up on filter quality in robust models on \textit{ImageNet1k} and noticed that robust models form more diverse filters than their non-robust counterparts. The presented study builds on top of this previous work, and explores differences specifically focused at the robustness aspect and with significantly more details. Instead of comparing robust models to a large collection of various \textit{ImageNet1k}-classifiers, we compare differences to the same architectures trained without robustness regularization to allow for a less biased analysis. Additionally, we extend our analysis to other datasets, refine previous quality metrics, and evaluate a new metric to capture the orthogonality of filterbanks. Regarding robust filter analysis, we observe that models trained for $\ell_\infty$ robustness form thresholding filters in early vision which are able to remove perturbations.\\

\paragraph{Connection to other model analysis}
\cite{frankle2018lottery} presented the \textit{Lottery Ticket Hypothesis}, which claims that CNNs form various redundant subnetworks that each increase the odds of finding a solution. Once a solution was found, the ``loosing'' subnetworks can be removed without any significant impacts on accuracy. Upon that, \cite{bai2021improving} observed that adversarial samples activate channels of the feature extractor more uniformly, and with larger magnitudes, and propose to suppress channels to increase robustness. Instead of suppressing channels, \cite{guo2022improving} showed that enhancing subnetworks can boost robustness to adversarial perturbations. \cite{utrera2021adversariallytrained} argue that adversarially-trained networks transfer better as they form richer representations.
We hypothesize that these findings correlate with filter quality and believe that degenerated filters are the ``loosing'' subnetworks, that are activated by adversarial attacks. Improving the filter quality, in theory, should be a necessary, but not sufficient criterion to achieve robustness.

%-------------------------------------------------------------------------
\section{Methods}
\begin{figure}
        \centering
        \begin{subfigure}{\columnwidth}
        \centering
            \includegraphics[width=0.9\linewidth]{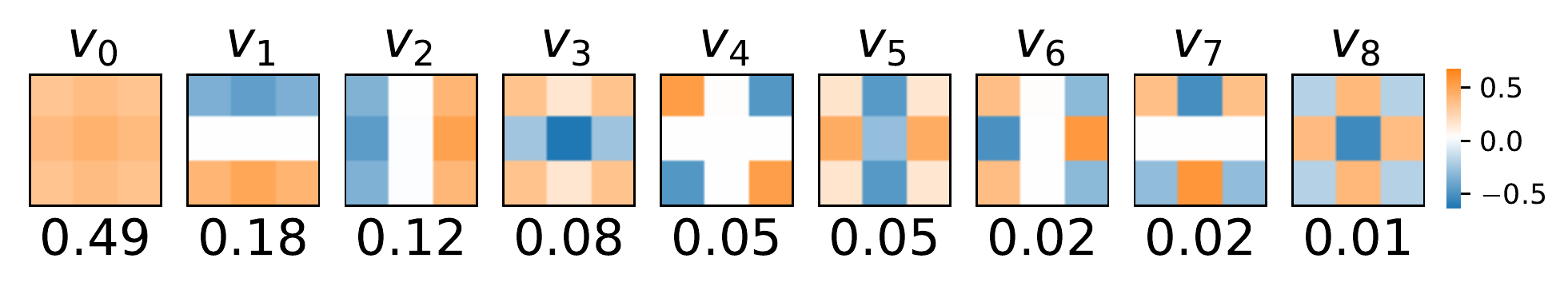} 
            \caption{All filters.}
            \label{fig:pca_basis_all}
        \end{subfigure}
        \begin{subfigure}{\columnwidth}
        \centering
            \includegraphics[width=0.9\linewidth]{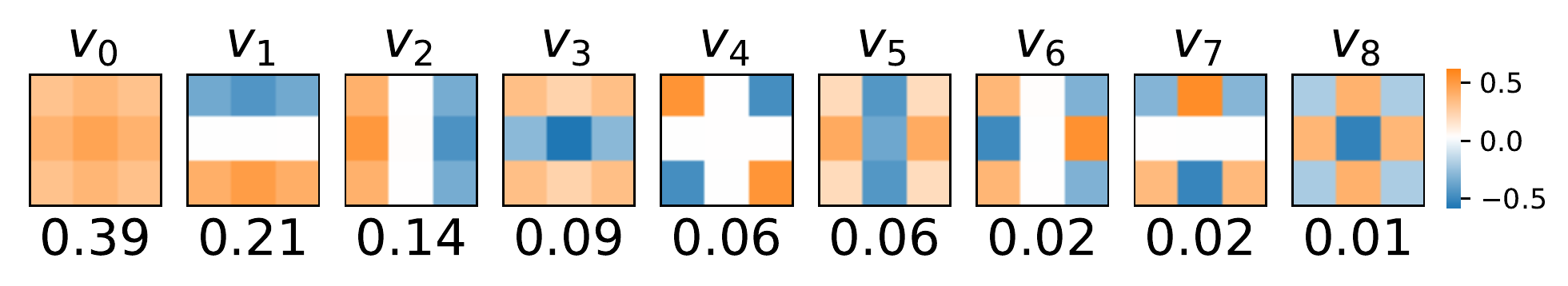}
            \caption{Filters from robust models.}
            \label{fig:pca_basis_robust}
        \end{subfigure}
        \begin{subfigure}{\columnwidth}
        \centering
            \includegraphics[width=0.9\linewidth]{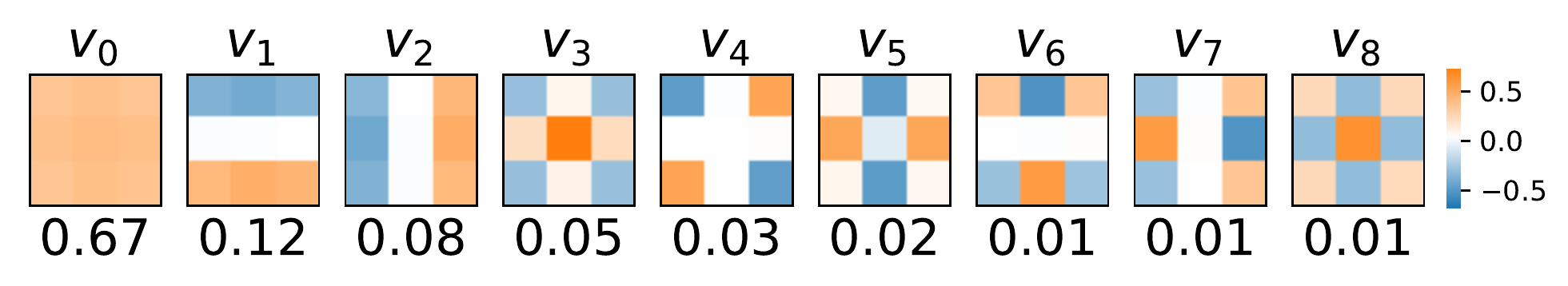}
            \caption{Filters from normal models.}
            \label{fig:pca_basis_normal}
        \end{subfigure}
    \caption{Filter basis and (cumulative) explained variance ratio per component (below) for filters from \protect\subref{fig:pca_basis_all} all models, \protect\subref{fig:pca_basis_robust} adversarially-robust models, \protect\subref{fig:pca_basis_normal} normal models. Basis vectors are sorted by decreasing variance.}
    \label{fig:pca_basis}
\end{figure}
For the following sections, let $\boldsymbol {W}^{(i)}\in\mathbb{R}^{c_{\text{out}}\times c_{\text{in}}\times k_{1} \times k_{2}}$ be the layer weight of the $i$-th convolution layer with $c_{\text{in}}$ input-channels, $c_{\text{out}}$ output-channels, and $\boldsymbol {F}\in \boldsymbol {W}^{(i)}$ being a filter with shape $k_1 \times k_2$ (here: $k_1=k_2=3$). For the following methods we reshape the layer weights into matrices that represent stacks of $n = c_{\text{out}}\times c_{\text{in}}$ flattened convolution filters:
\begin{equation}
\boldsymbol {W}^{(i)}\in\mathbb{R}^{c_{\text{out}}\times c_{\text{in}}\times k_{1} \times k_{2}} \rightarrow \boldsymbol {W}^{(i)}\in\mathbb{R}^{n\times (k_{1} \cdot k_{2})}    
\end{equation}

\paragraph{Comparing filter structure} 
As in \cite{anonCVPRPaper} we perform a principal component analysis (PCA) via singular-value decomposition (SVD)\cite{Jolliffe1986} to understand the filter structure. In this work, however, we aim at reducing the previously hefty impact of sparse filters, by removing them from layer weights (for details see the next paragraph). Then we normalize each filter $\boldsymbol {F}\in \boldsymbol {W}^{(i)}$ individually to $\boldsymbol {F}'$:
\begin{equation}
    \begin{split}
        \displaystyle d_i &= \max_{i,j} \left|\boldsymbol {F}_{ij}\right|\\
        \boldsymbol {F}_{ij}' &= \left. 
        \begin{cases}
            \boldsymbol {F}_{ij} / d_i, & \text{if } d_i \neq 0 \\
            \boldsymbol {F}_{ij}, & \text{else} 
      \end{cases}
      \right.\\
    \end{split}
\end{equation}
The resulting layer weight matrix is centered and decomposed via SVD into a ${n\times k_{1} \cdot k_{2}}$ rotation matrix $\boldsymbol {U}$, a $k_{1} \cdot k_{2}\times k_{1} \cdot k_{2}$ diagonal scaling matrix $\boldsymbol {\Sigma}$, and a $k_{1} \cdot k_{2}\times k_{1} \cdot k_{2}$ rotation matrix $\boldsymbol {V}^{T}$. The diagonal entries $\sigma_{i}, i=0,\dotsc ,n-1$ of $\boldsymbol {\Sigma}$ form the singular values in decreasing order of their explained variance. The row vectors $v_{i}, i=0,\dotsc ,k_{1} \cdot k_{2}-1$ in $\boldsymbol {V}^{T}$ are called principal components/basis vectors. Every row vector $c_{ij}, j=0,\dotsc ,k_{1} \cdot k_{2}-1$ in $\boldsymbol {U}$ is the coefficient vector for $F'_{i}$.%
\begin{equation}
    \begin{split}
        \boldsymbol {W} - \bar{\boldsymbol {W}} &= \boldsymbol {U}\boldsymbol {\Sigma} \boldsymbol {V}^{T} \\
    \end{split}
\end{equation}
Where $\bar{\boldsymbol {W}}$ denotes the vector of column-wise mean values of any matrix $\boldsymbol {W}$. We can then measure the explained variance ratio $\hat{a}$ of each principal component.%
\begin{equation}
    \begin{split}
        {a} &= {(\boldsymbol {\Sigma}\boldsymbol {I})^{2}}/{(n - 1)} \\
        \hat{a} &= {{a}}/{\|{a}\|_{1}} 
    \end{split}
\end{equation}
The sum of principal components $v_{i}$ weighted by the coefficients $c_{i}$ allows to reconstruct every scaled filter $\boldsymbol {F'} \in \boldsymbol {W}^{(i)}$.
\begin{equation}
    \begin{split}
        \displaystyle \boldsymbol {F}'&=\sum_{i}c_{i}v_{i} + \bar{\boldsymbol {W}}_{i}
    \end{split}
\end{equation}

\paragraph{Measuring layer quality}
A static measurement of the layer quality (meaning by only using learned parameters) can be obtained by measuring the ratio of filters where all weights are near-zero (\textit{sparsity}) as these filters do not contribute to the feature maps due to their low magnitudes. We apply the criterion presented in \cite{anonCVPRPaper} and call a filter $\boldsymbol {F} \in \boldsymbol {W}^{(i)}$ sparse if $\max |\boldsymbol {F}| \leq \max |\boldsymbol {W}^{(i)}|/100$.
The ratio is then defined by:
\begin{equation}
    \begin{split}
        |\{\boldsymbol {F} | \boldsymbol {F} \in \boldsymbol {W}^{(i)} \land (\forall x \in \boldsymbol {F}: -\epsilon_{0} \leq x \leq \epsilon_{0})\}|/ n
    \end{split}
\end{equation}
Additionally, it is possible to quantify the diversity of filter structure in a given layer. We fit the PCA to individual layer weights $\boldsymbol {W}^{(i)}$ without normalization of filters and again remove sparse filters. The diversity can be then estimated by the non-negative $log_{10}$ entropy of the explained variance of each basis vector $H(\hat{a})$ (\textit{variance entropy}) \cite{anonCVPRPaper}.%
\begin{equation}
    H(\mathcal{X})=\sum _{x\in\mathcal{X}}{x\log_{10} x}
\end{equation}
An entropy variance value of $H([1, 0, \dotsm, 0])=0$ indicates a homogeneity of present filters, while the maximum $H(\mathbbm{1})$ (here: $0.954$) indicates a uniformly spread variance across all basis vectors, as found in random, non-initialized layers \cite{anonCVPRPaper}. Values close to both edges indicate a degeneration. 

Additionally, \textit{orthogonality} is a desirable property in convolutional weights \cite{brock2017neural, brock2018large}, as it helps with gradient propagation and is directly coupled with diversity of generated feature maps. Due to computational limits we measure the orthogonality between filterbanks (i.e. stacks of $c_{\text{in}}$ filters) instead of individual filters. The filterbanks are normalized to unit-length.
\begin{equation}
    \begin{gathered}
    1 - \frac{
        \|\boldsymbol {W}^{(i)}(\boldsymbol {W}^{(i)})^{T} - \boldsymbol {I}\|_{1}
    }{c_{\text{out}}\cdot(c_{\text{out}}-1)}
    \end{gathered}
\end{equation}
An orthogonality value of 1 stipulates the orthogonality of all filterbanks in a layer, whereas 0 indicates parallel filterbanks that produce perhaps differently scaled but otherwise identical feature maps.

\paragraph{Quantifying distribution shifts}
We quantify distribution shifts between two distributions $P, Q$ by a symmetric, non-negative variant of KL-divergence \cite{kullback1951}. For multi-dimensional distributions we compute the divergence on each axis $i$ and sum it weighted them by a factor $w_i$:
\begin{equation}
    \begin{split}
        \displaystyle\sum_{i}{w}_{i}\sum_{x\in {\mathcal {X}}}P_i(x)\log {\frac {P_i(x)}{Q_i(x)}}+Q_i(x)\log {\frac {Q_i(x)}{P_i(x)}}
        \end{split}
\end{equation}

\paragraph{Models}
We collected 71 robust model checkpoints \cite{rebuffi2021fixing,huang2022exploring,zhang2020attacks,zhang2019propagate,hendrycks2019using,zhang2021geometryaware,chen2021ltd,andriushchenko2020understanding,cui2021learnable,rice2020overfitting,dai2021parameterizing,gowal2021uncovering,sitawarin2021sat,chen2021efficient,zhang2019theoretically,wu2020adversarial,wong2020fast,huang2020selfadaptive,carmon2022unlabeled,pang2020boosting,gowal2021improving,sehwag2020hydra,sridhar2021improving,chen2020adversarial,sehwag2021robust,addepalli2021towards,Ding2020MMA,rade2021helperbased,Wang2020Improving,robustness} from the $\ell_\infty$-\textit{RobustBench} leaderboard \cite{croce2021robustbench}.
Additionally, for each appearing architecture we trained an individual model, without any specific robustness regularization, and without any external data (even if the robust counterpart relied on it). 
% \textit{CIFAR-10/100} data was zero-padded by 4 px along each dimension, and then transformed using a $32^2$ px random crops, and random horizontal flips. For the hyper parameters an initial learning rate of 1e-8, a weight decay of 1e-2, a batch-size of 256 and a Nesterov-momentum of 0.9 is used. The employed SGD optimizer is scheduled to decrease the learning rate every 30 epochs by a factor of $\gamma = 0.1$. The models are trained for 125 epochs. The loss is determined using Categorical Cross Entropy. 
Training \textit{ImageNet1k} architectures with these parameters resulted in rather poor performance and we replaced these models with pretrained \textit{ImageNet1k} models included from the \textit{timm}-library \cite{rw2019timm}.
\begin{table}
    \centering
    \small
    \begin{tabular}{p{1.1cm}p{1.25cm}p{1.25cm}p{1.25cm}p{1.25cm}}
        \toprule
        %{} &  Adv. Trained Clean Acc &  Adv. Trained Robust Acc &  Norm. Trained Clean Acc \\
        %Dataset  &                         &                          &                          \\
        {} & {} & \multicolumn{1}{c}{Normal} & \multicolumn{2}{c}{Robust}\\
        \cmidrule{4-5}
        {Dataset} & {$3\times 3$ Filters [M]} & {Clean Acc.} & {Clean Acc.} & {Robust Acc.}\\
        \midrule
        cifar10  & $9.1\pm 10.5$ &             $92.2\pm 4.2$ &          $86.9\pm 2.6$ &             $56.7\pm 5.9$ \\
        cifar100 & $9.3\pm 10.6$ &             $72.7\pm 8.5$ &         $62.2\pm 3.9$ &             $29.0\pm 3.9$ \\
        imnet & $2.0\pm 1.7$ &             $78.5\pm 4.9$ &           $60.7\pm 6.3$ &             $30.8\pm 5.6$ \\
        \bottomrule
    \end{tabular}
    \caption{Comparison between average (and \textit{std}) performance and parameter size on all evaluated datasets. Clean Acc. refers to the regular validation accuracy, while Robust Acc. refers to the robust accuracy as measured by \textit{RobustBench}.}
    \label{tab:model_performance}
\end{table}
%-------------------------------------------------------------------------
%
\section{Results}
\subsection{Filter basis} 
In a first step, we investigate the basis forming the obtained filters. We therefore separate the filters extracted from all models into three filter sets: all filters, only filters from robust models, and only filters from normal models. Then we apply the filter structure measurement to each set individually.

We observe that the basis-vectors obtained from all three sets do not significantly differ (\cref{fig:pca_basis}). Changes only include minor fluctuations (note that basis vectors can be inverted which is equivalent with inverting the coefficients). However, while $67\%$ of the normal filter variance can be reconstructed from the first basis vector alone, robust models show a more uniform distribution of the variance, suggesting that these models form more structurally diverse filters.
%-------------------------------------------------------------------------
%
\begin{figure*}
  \centering
  \includegraphics[width=\linewidth]{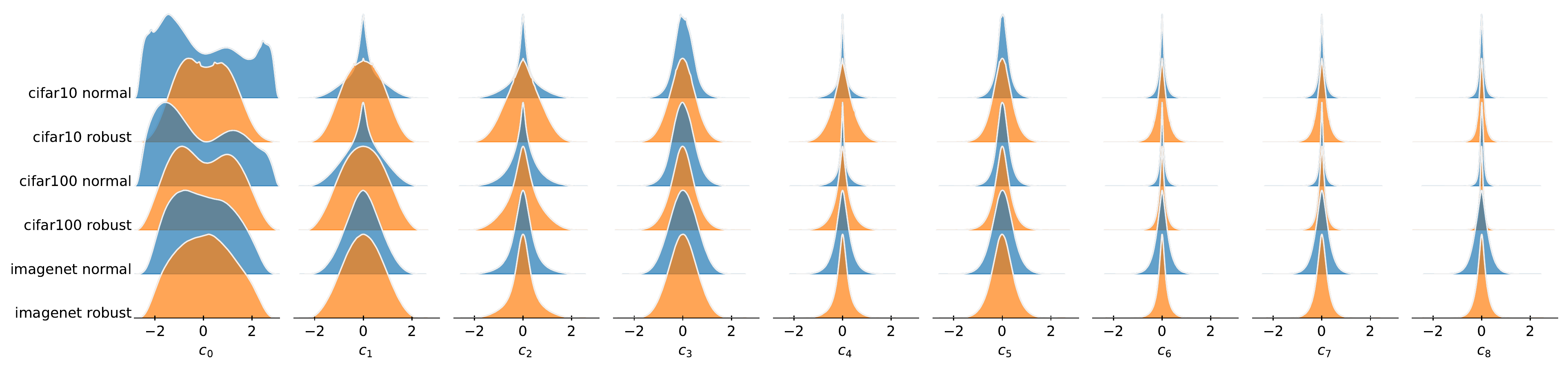}
  \caption{Coefficient distribution along every basis of robust (adversarially-trained) vs. normal models on different datasets. Shifts between robust and normal models appear to decrease with dataset complexity.}
  \label{fig:ridge_dataset_by_robust}
\end{figure*}
\subsection{Filter structure}
In this section we aim at understanding the differences in the filter structure. For this, we compute a common basis consisting of all collected filters and measure shifts between coefficients separated by dataset and regularization. We weigh the distributions by the explained variance ratio of the respective axis.\\

\paragraph{Coefficient shifts by dataset}
The coefficient distributions (\cref{fig:ridge_dataset_by_robust}) show clear shifts between robust and normal models, but this shift decreases with increasing complexity of the dataset. We obtain a weighted KL-divergence of 0.55, 0.16, and 0.01 for \textit{CIFAR-10/100}, and \textit{ImageNet1k} respectively. Interestingly, we also see a reduced drift for \textit{CIFAR-100} compared to \textit{CIFAR-10}, although it has the same amount of training samples but more classes. This suggests that more complex datasets lead to smaller distribution shifts between robust and normal models, with an emphasis on the fact that complexity does not only refer to the amount of input training data.
It is worth noting that robust models achieve a significantly worse clean accuracy than their counterparts and this performance gap increases with dataset complexity (\cref{tab:model_performance}). On average, robust accuracy is an additionally 30\% worse for all studied datasets. And \textit{ImageNet1k} models are evaluated with a different $\epsilon$, which may hide their true (non)-robustness. Additionally, the studied \textit{ImageNet1k} models on average only employ 2M $3\times 3$ filters (plus a negligible amount of larger filters), while the models on the arguably simpler datasets employ 9M on average. It is therefore likely, that \textit{CIFAR-10/100} shows an increased effect of degeneration, and, that \textit{ImageNet1k} pairs similarity is due to smaller architectures, and lesser robustness performance, rather than intrinsic similarity.\\
%It is therefore unclear if a simple inclusion of additional data or a surrogate task would reduce the distribution shift between adversarially- and normally trained models.
% This is inline with findings that surrogate tasks \todo{cite} increase robustness by putting an additional strain on trained models.
%

\paragraph{Coefficient shifts by depth}
Following the previous observation, we investigate the most significant shifts in filter coefficients and measure the divergence at various stages of depth. To compare models with different depths, we group filter coefficients in deciles of their relative depth in the model. The obtained shifts (\cref{fig:divergence_by_depth}) seem to increase with convolution depth and peak in the last 20\% of the depth for \textit{CIFAR-10/100}. For \textit{ImageNet1k} the peak shift is measured in the 8th decile, whereas the shift in later stages is minimal. Aside of the shifts in later stages, for all datasets, there is relatively low shift throughout the depth with the most salient outlier being seen in the very first convolution layer.\footnote{The primary convolution stage in \textit{ImageNet1k}-models use a larger kernel-size and is therefore not included in this study, yet we expect similar observations there.} This outlier is indeed limited to the first layer, adding filters from the secondary layers vanishes the shift. Once again the maximum shift appears to decrease with dataset complexity.
\begin{figure}
    \centering
    \begin{subfigure}{\columnwidth}
        \includegraphics[width=\linewidth]{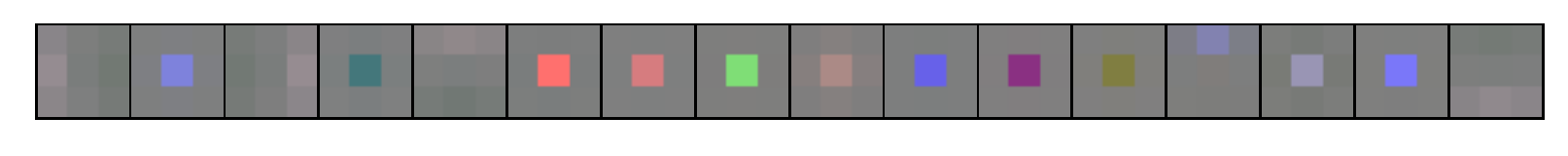}
        \caption{Adversarial training.}\label{fig:addepalli2021towards_rgb_filters_robust}
    \end{subfigure}
    \begin{subfigure}{\columnwidth}
        \includegraphics[width=\linewidth]{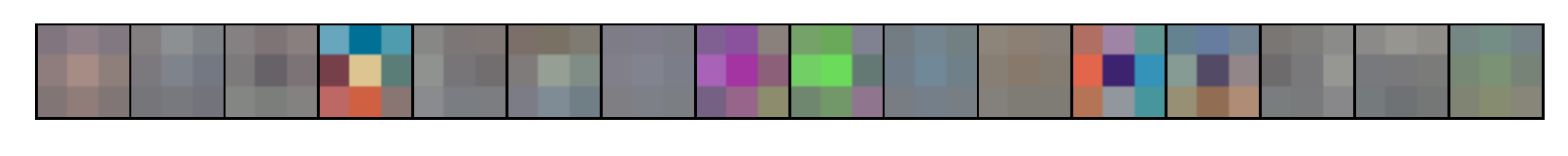}
        \caption{Normal training.}
        \label{fig:addepalli2021towards_rgb_filters_normal}
    \end{subfigure}
    \caption{Full set of first stage convolution filters of a \textit{WideResNet-34-10} trained with \protect\subref{fig:addepalli2021towards_rgb_filters_robust} adversarial training as provided by \cite{addepalli2021towards} on \textit{CIFAR-10} and \protect\subref{fig:addepalli2021towards_rgb_filters_normal} normal training.}
    \label{fig:rgb_filters}
\end{figure}
\begin{figure}
    \centering
   \begin{subfigure}{\columnwidth}
        \includegraphics[width=\linewidth]{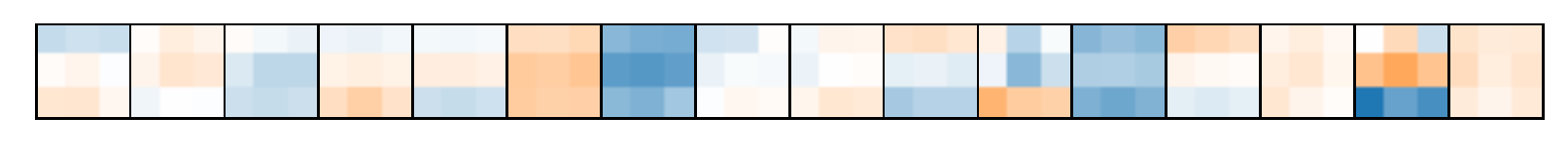}
        \caption{Adversarial training.}\label{fig:last_filters_robust}
    \end{subfigure}
    \begin{subfigure}{\columnwidth}
        \includegraphics[width=\linewidth]{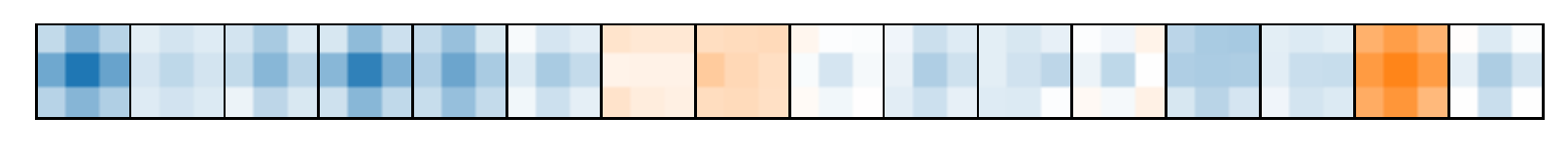}
        \caption{Normal training.}
        \label{fig:last_filters_normal}
    \end{subfigure}
    \caption{Randomly selected convolution filters of the last convolution layer in a \textit{WideResNet-34-10} trained with \protect\subref{fig:last_filters_robust} adversarial training as provided by \cite{addepalli2021towards} on \textit{CIFAR-10} and \protect\subref{fig:last_filters_normal} normal training.}
    \label{fig:last_filters}
\end{figure}

\begin{figure}
  \centering
  \includegraphics[width=\linewidth]{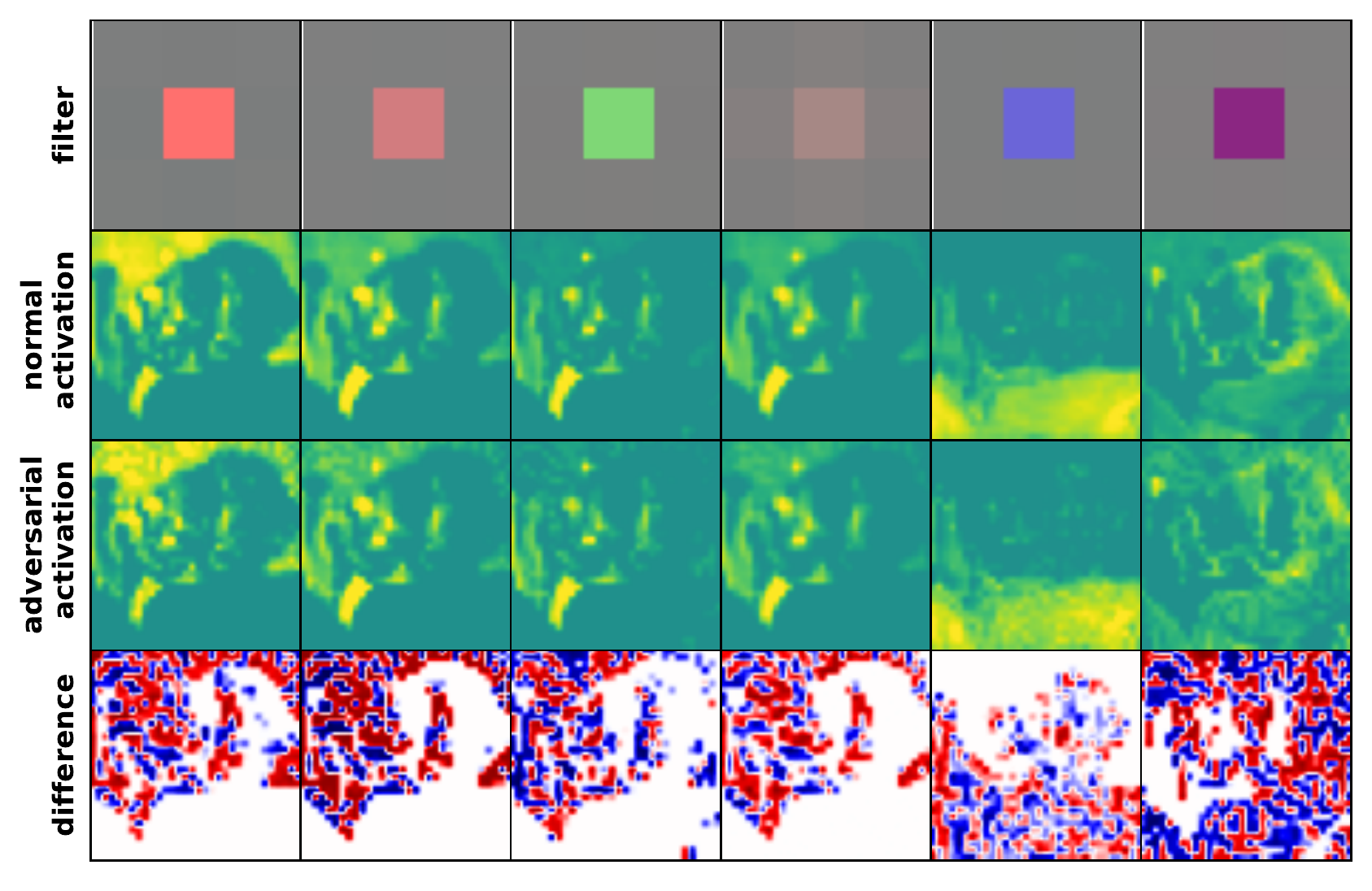}
  \caption{Activations generated by (randomly selected) thresholding-filterbanks (top) in the primary convolution stage of a robust \textit{WideResNet-34-10} by \cite{addepalli2021towards}. The first row shows the thresholding-filters. The second row shows the activations of each filter for an input sample, and the same sample with perturbations, respectively. Finally, the last row shows the difference in activations: perturbations (red, blue) are removed from ROIs (white).}
  \label{fig:primary_thresholding}
\end{figure}

\paragraph{First and last convolution layer}
To better understand the cause of the observed distribution shifts we visualize the first and last convolution layers.
The primary convolution stage (\cref{fig:rgb_filters}) shows a striking difference: Normal models show an expected \cite{Yosinski2014} diverse set of various filters, yet, almost all robust models develop a large presence of filters performing a weighted summation of the input channels (as $1\times 1$ convolutions would do). We hypothesize that in combination with the common \textit{ReLU}-activations (and their derivatives) these filters perform a thresholding of the input data which can eliminate small perturbations. Indeed, plotting the difference in activations for natural and perturbed samples ( \cref{fig:primary_thresholding}), allows us to obtain visual confirmation that these filters are successful in removing perturbations from various regions of interest (ROI), \eg from the cat, background, foreground.
For the deepest convolution layers (\cref{fig:last_filters}) we observe the opposite: normal filters show a clear lack of diversity, and mostly remind of gaussian blur filters, while adversarially-trained filters appear to be richer in structure and are more likely to perform complex transformations. Contrary to the distinct primary layer, this observation is visible across multiple deeper layers.

%%%%%%%
\subsection{Filter quality}
\begin{figure*}
  \centering
  \includegraphics[width=\linewidth]{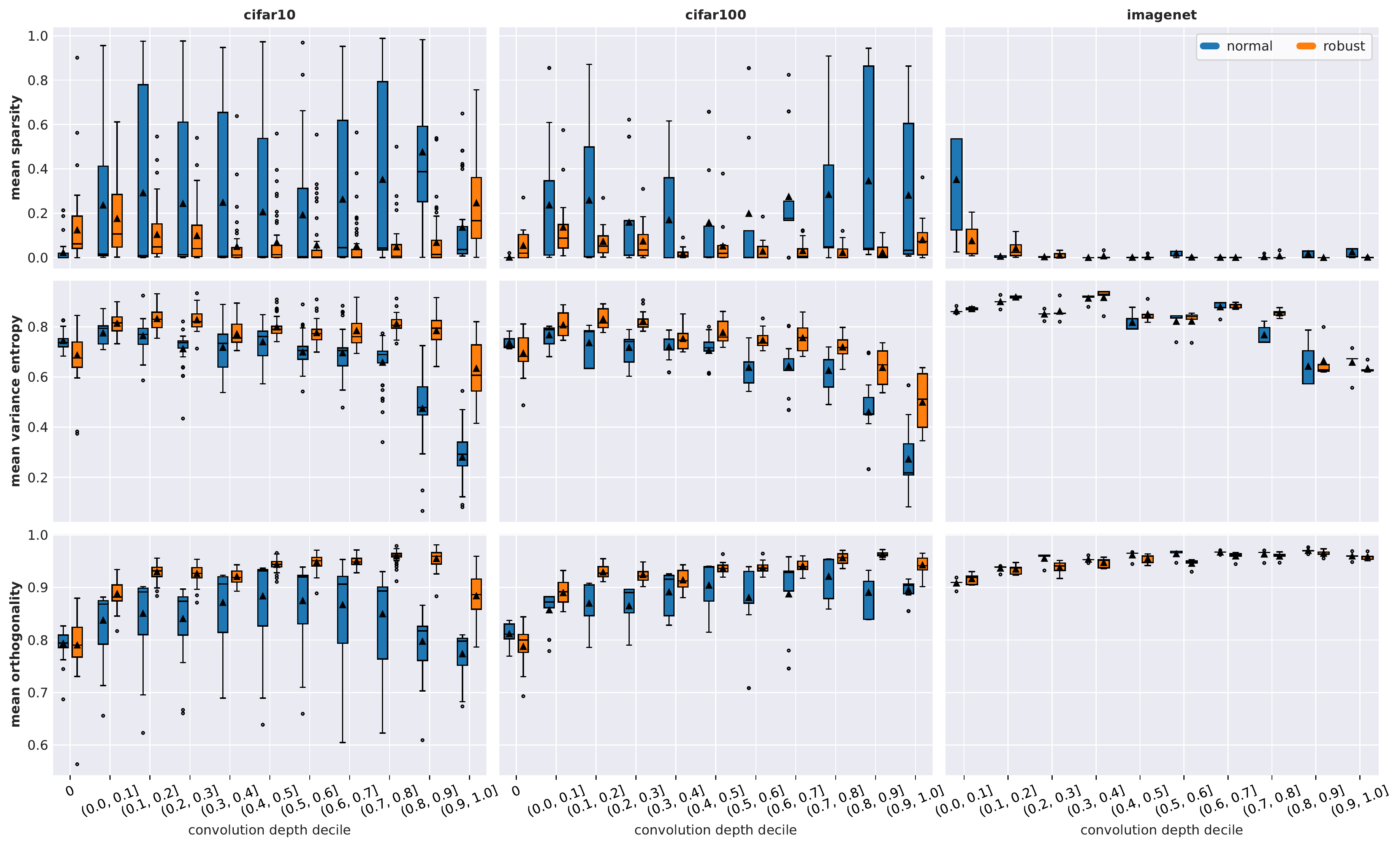}
      
  \caption{Distribution of filter quality comparison by depth measured via \textit{sparsity} (top), \textit{variance entropy} (center), and \textit{orthogonality} (bottom) between normal and adversarial-training for \textit{CIFAR-10} (left), \textit{CIFAR-100} (center), \textit{ImageNet1k} (right) datasets.}
  \label{fig:metrics_normal_vs_robust}
\end{figure*}
While the prevenient analysis focused on distribution shifts in filter structure this section focuses on the related quality aspect of filters. In particular, we measure the amount of contributing filters through \textit{sparsity}; the diversity of filters through \textit{variance entropy}; and the redundancy of filterbanks through \textit{orthogonality}. Similarly to the findings in structure, we observe fewer differences in quality with dataset complexity (\cref{fig:metrics_normal_vs_robust}), but also a general increase in quality for both robust and normal models. The results on \textit{ImageNet1k} are less conclusive due to a near-optimal baseline and a low sample size.\\

\paragraph{Sparsity} We observe a very high span of sparsity across all layers for normal models that decreases with dataset complexity. Robust training significantly further minimizes sparsity and it's span across all depths.  Notable outliers include the primary stages, as well as the deepest convolution layers for \textit{CIFAR-10}. Generally, sparsity seems to be lower in middle-stages.\\

\paragraph{Variance entropy} The average variance entropy is relatively constant throughout the model but decreases with deeper layers. The entropy of robust models starts to decrease later and less significantly but the difference between diminishes with dataset complexity. Compared to \textit{CIFAR-10}, robust \textit{CIFAR-100} models shows a lower entropy in deeper layers, while there is no clear difference between normal models. \textit{ImageNet1k} models show a higher entropy across all depths. \\

\paragraph{Orthogonality} Robust models show an almost monotonic increase in orthogonality with depth, except for the last decile, whereas normal models eventually begin to decrease in orthogonality. Again the differences diminish with dataset complexity and the span in obtained measurements of non-robust models is crucially increased.

%-------------------------------------------------------------------------
\section{Conclusion}

Adversarially-trained models appear to learn a particularly more diverse, less redundant, and less sparse set of convolution filters than their non-regularized variants do. We assume that the increase in quality is a response to the additional training strain, as the more challenging adversarial problem occupies more of the available model capacity that would otherwise be degenerated. We observe a similar effect during normal training with increasing dataset complexity. However, although the filter quality of normally trained \textit{ImageNet1k} models is exceptionally high, their robustness is not. So, filter quality alone is not a sufficient criterion to establish robustness. We end with the following currently unanswered questions: \textit{Is dataset complexity the cause for lower quality shifts, or, is the difference we measure merely a side effect of heavily overparameterized architectures in which adversarial training can close the gap to more complex datasets? If not, can we increase robustness by filter quality regularization during training?}

%%%%%%%%% REFERENCES
\clearpage
{\small
\bibliographystyle{ieeetr_fullname}
\bibliography{egbib, main, distill, robustmodels}
}

%%%%%%%%%%%%%%%%%%%%%%
%APPENDIX NOT ALLOWED%
%%%%%%%%%%%%%%%%%%%%%%

\end{document}